\documentclass[10pt,twocolumn,letterpaper]{article}
\usepackage{cvpr}
\usepackage{times}
\usepackage{epsfig}
\usepackage{graphicx}
\usepackage{amsmath}
\usepackage{amssymb}
\usepackage{multirow}
\usepackage{caption}
\usepackage{accents}
\usepackage{xcolor}
\usepackage{subcaption}
\usepackage{soul}

\usepackage[pagebackref=true,breaklinks=true,letterpaper=true,colorlinks,bookmarks=false]{hyperref}
\cvprfinalcopy

\newcommand{\customfootnotetext}[2]{{
\renewcommand{\thefootnote}{#1}
\footnotetext[0]{#2}}}

\def\cvprPaperID{206} 

\def\sexyname{Cops-Ref\xspace}

\ifcvprfinal\fi
\begin{document}

\title{Cops-Ref: A new Dataset and Task on Compositional Referring \\Expression Comprehension}

\author{
Zhenfang Chen$^1$\thanks{Work done while Zhenfang Chen was visiting the University of Adelaide.}\quad
Peng Wang$^2$\quad
Lin Ma$^{3}$\quad
Kwan-Yee K. Wong$^1$\quad
Qi Wu$^4$\thanks{Corresponding author.}\\
$^1$The University of Hong Kong  \quad $^2$University of Wollongong \\
$^3$Tencent AI Lab  \quad  $^4$Australian Centre for Robotic Vision, University of Adelaide\\
\texttt{\small{$^1$\{zfchen, kykwong\}@cs.hku.hk}} \quad \texttt{\small{$^2$pengw@uow.edu.au}} \\
\texttt{\small{$^3$forest.linma@gmail.com}}  \quad \texttt{\small{$^4$qi.wu01@adelaide.edu.au}}
}
\twocolumn[{%
\renewcommand\twocolumn[1][]{#1}%
 \maketitle
 \vspace{-2em}
    \begin{tabular}{l}
    {\small\textbf{Reasoning tree}: cat (left, sleeping) $\xrightarrow{\text{resting}}$ towel (white) }\\
    \small\textbf{Expression}: \textit{The cat on the left that is sleeping and resting on the white towel.}
    \end{tabular}
    \vfill
    \centering
    \begin{minipage}[c]{0.18\textwidth}
    \centering
    \includegraphics[width=\textwidth]{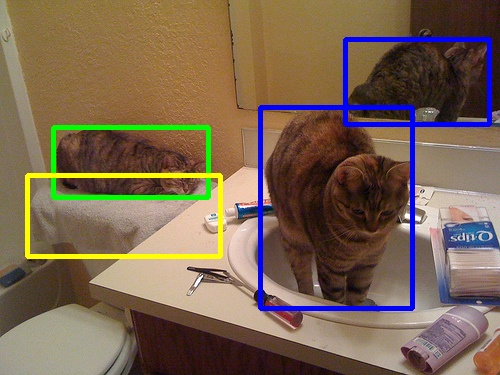}
    \scriptsize{(a) The image with the target ``cat''}
    \end{minipage}
    \begin{minipage}[c]{0.36\textwidth}
    \centering
    \includegraphics[width=\textwidth]{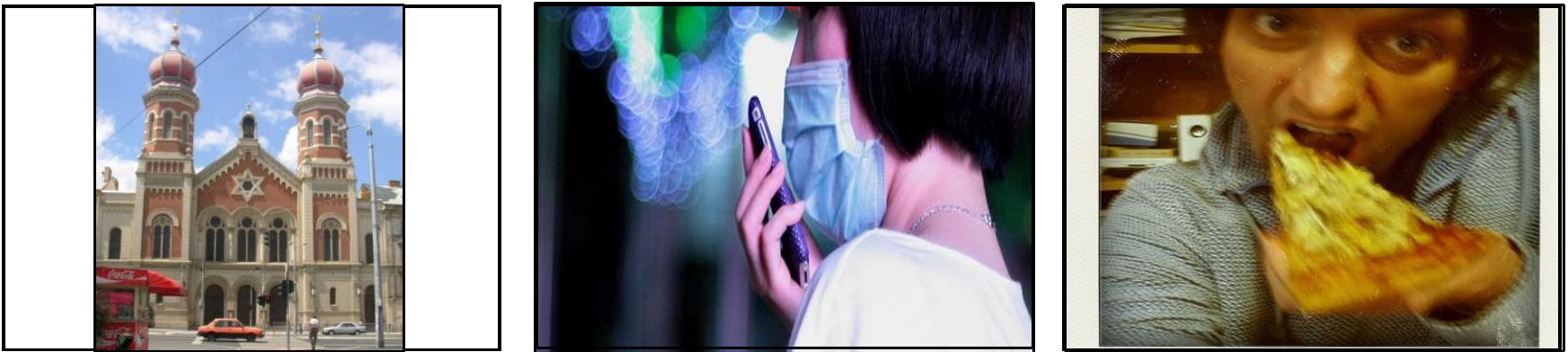}
    \scriptsize{(b) Distractors of different categories}
    \includegraphics[width=\textwidth]{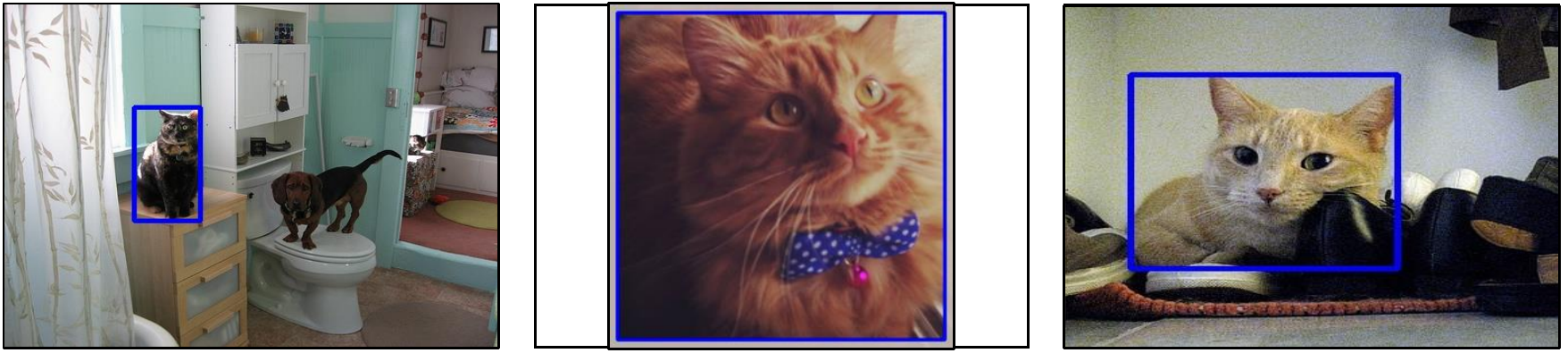}
    \scriptsize{(c) Distractors with ``cat''}
    \end{minipage}
    \begin{minipage}[c]{0.36\textwidth}
    \centering
    \includegraphics[width=\textwidth]{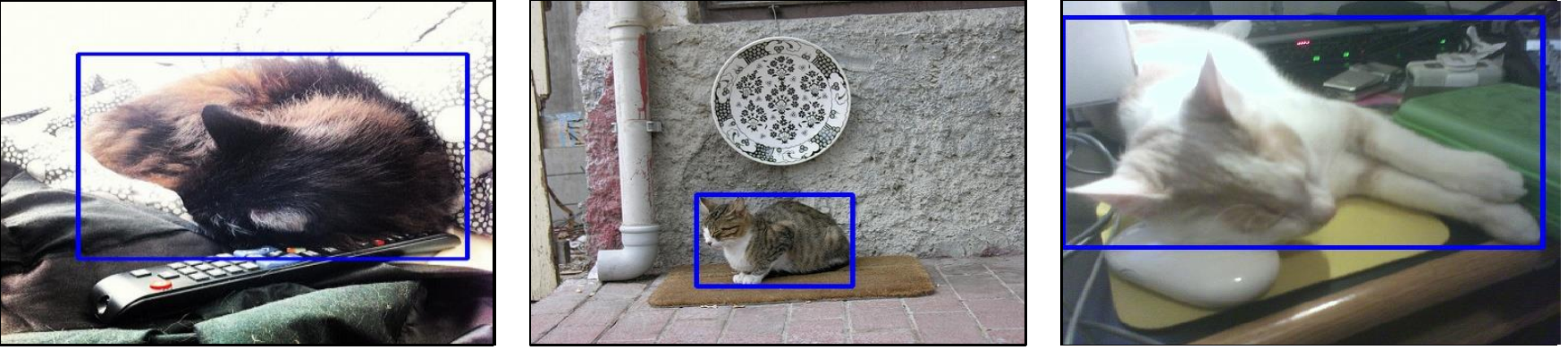}
    \scriptsize{(d) Distractors with ``sleeping cat''}
    \includegraphics[width=\textwidth]{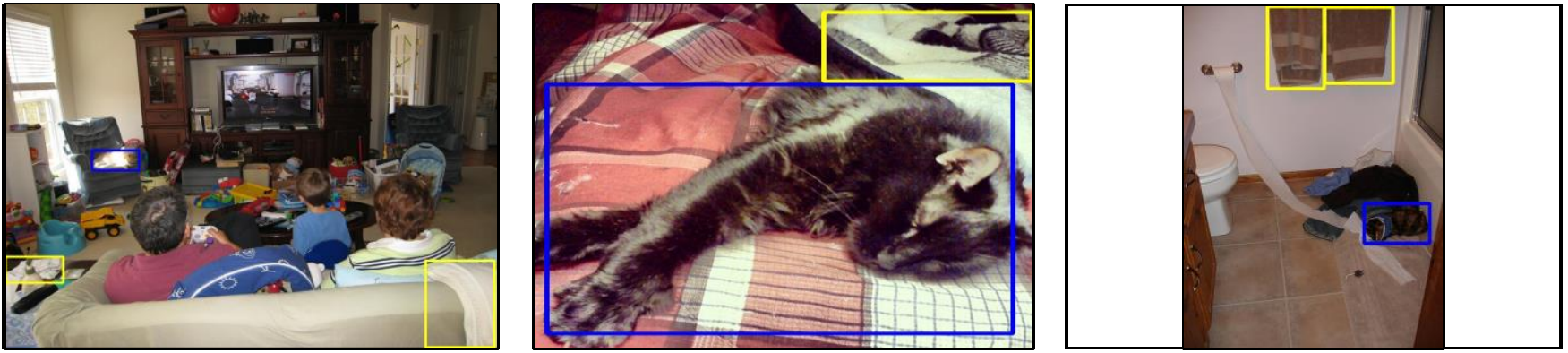}
    \scriptsize{(e) Distractors with ``cat'' and ``towel''}
    \end{minipage}
    \vspace{-0.7em}
    \captionof{figure}{An example from the new Cops-Ref dataset for compositional referring expression comprehension. The task requires a model to identify a target object described by a compositional referring expression from a set of images including not only the target image but also some other images with varying distracting factors as well. The target/related/distracting regions are marked by green/yellow/blue boxes, respectively. More details about the reasoning tree can be seen in Sec. \ref{sub:expEngine}.}\label{fig:task}
    \vspace{0.7em}
}]
\customfootnotetext{*}{Work done while Zhenfang was visiting the University of Adelaide.}
\customfootnotetext{$\dagger$}{Corresponding author.}
\begin{abstract}
    Referring expression comprehension (REF) aims at identifying a particular object in a scene by a natural language expression. It requires joint reasoning over the textual and visual domains to solve the problem. Some popular referring expression datasets, however, fail to provide an ideal test bed for evaluating the reasoning ability of the models, mainly because 1) their expressions typically describe only some simple distinctive properties of the object and 2) their images contain limited distracting information. 
    To bridge the gap, we propose a new dataset for visual reasoning in context of referring expression comprehension with two main features. First, we design a novel expression engine rendering various reasoning logics that can be flexibly combined with rich visual properties to generate expressions with varying compositionality. Second, to better exploit the full reasoning chain embodied in an expression, we propose a new test setting by adding additional distracting images containing objects sharing similar properties with the referent, thus minimising the success rate of reasoning-free cross-domain alignment. We evaluate several state-of-the-art REF models, but find none of them can achieve promising performance. A proposed modular hard mining strategy performs the best but still leaves substantial room for improvement. {We hope this new dataset and task can serve as a benchmark for deeper visual reasoning analysis and foster the research on referring expression comprehension.}

\end{abstract}

\vspace{-2em}
\section{Introduction}
In recent years, computer vision tasks that require high-level reasoning have attracted substantial interest. Visual question answering (VQA)~\cite{hudson2019gqa,balanced_binary_vqa} and visual dialog (VD)~\cite{visdial,Kottur_2018_ECCV} are typical examples of such a trend, where the system answers free-form questions based on an image by jointly reasoning over the textual and visual domains. A prerequisite to achieve this ultimate goal of artificial intelligence is the ability to ground the rich linguistic elements embodied in the language onto the visual content of the image. Referring expression comprehension (REF) is such a visual grounding task, which targets at identifying a particular object in a scene by an expression phased in natural language.
A number of datasets~\cite{kazemzadeh2014referitgame,mao2016generation,yu2016modeling}  have been constructed for this task, and on top of which various models~\cite{liu2019improving, rohrbach2016grounding, yu2018mattnet} have been developed.

Such popular datasets~\cite{kazemzadeh2014referitgame,mao2016generation,yu2016modeling}, however, cannot serve as ideal test beds to evaluate the reasoning ability of REF models. First, the expressions are typically simple and short, focusing mainly on some distinctive properties of the referent, such as object categories, attributes, or some simple relationships. For example, only some superficial reasoning is involved in expressions like `\texttt{the girl with glasses}' and `\texttt{the man sitting next to a table}'. Secondly, many images in the existing datasets contain only limited distracting information (\eg, containing only two or three objects of the same category) and do not necessitate complex reasoning. As an example, although we are given a complex expression `\texttt{The cat on the left that is sleeping and resting on the white towel}'. to localise the target cat in the example image shown in Fig.~\ref{fig:task}~(a), we can still have high chance to succeed even if we only use a simple expression `\texttt{The cat on the left}' as the query.
Another non-negligible issue is dataset bias.
As stated by Cirik \etal~\cite{cirik2018visual}, a system that ignores the expression but uses only the image as input can still outperform random guess by a large margin. 
Recently, a synthetic dataset for referring expression, CLEVR-Ref+~\cite{liu2019clevr}, is proposed to facilitate the diagnosis of visual reasoning. However, this dataset sacrifices visual realism and semantic richness by only describing some simple shapes and attributes.

To tackle the aforementioned problems, we propose a new challenging dataset for visual reasoning in context of referring expression comprehension. Our dataset is built on top of the real-world images in GQA~\cite{hudson2019gqa} and therefore it pertains visual realism and semantic richness. The key novelty of our dataset lies in a new expression engine. We design six reasoning logics, \ie \emph{and, or, order, same, not, chain}, which can be flexibly combined with the rich visual information (\eg, object categories, visual attributes, location information, and object interactions) to generate expressions with varying compositionality levels. Moreover, to overcome the sparse emergence of object categories and dataset bias, we design a new test setting by adding distracting images that contain objects sharing similar visual properties with the referent (\eg, same object category and similar attributes). Along with the dataset, a new REF task named COmPoSitional Referring expression comprehension (\sexyname) is proposed, requiring a model to localise a region described by the flowery expression from a set of visually similar images. With the new dataset and task, the success rate of reasoning-free cross-domain alignment can be minimised.

We evaluate various state-of-the-art REF models on our proposed \sexyname dataset, but we find none of them can achieve a  satisfactory performance. A modular hard-mining strategy is proposed to automatically mine hard negative examples embodying different visual properties. It achieves the best performance on the \sexyname task but still leaves much room for further improvement.

The contributions of this paper can be summarised as follows:
    1) We introduce a new challenging task named \sexyname,
    which requires a model to localise the referent from a set of images with objects sharing similar visual properties.
    2) We build a new dataset on top of real images, pertaining visual realism and semantic richness. It can complement the synthetic reasoning dataset to evaluate models' reasoning ability more rigorously.
    3) We design a novel expression engine. It supports various reasoning logics that can be flexibly combined with rich visual stimuli to generate expressions with varying compositionality.
    4) We conduct comprehensive evaluation on the REF models, among which the proposed modular hard mining strategy performs best but still leaves much room for improvement.

\vspace{-1em}
\section{Related Work}
\vspace{-0.5em}
\paragraph{Referring Expression Datasets.}
Toward tackling the REF task, a number of datasets~\cite{kazemzadeh2014referitgame, mao2016generation,plummer2015flickr30k,yu2016modeling,chen19acl} have been constructed by asking annotators to provide expressions describing regions of images. However, it is labor-intensive and hard to control the annotation quality and most of the queries in the datasets can be easily solved by simply reasoning on object categories, attributes and shallow relations.
Inspired by the synthetic dataset CLEVR~\cite{johnson2017clevr} for visual question answering (VQA), Liu~\etal~\cite{liu2019clevr} built a synthetic REF dataset, CLEVR-Ref+, by synthesising both images and expressions. However, it has been noticed in \cite{hudson2019gqa} that images in CLEVR, with only a handful of object classes, properties and spatial relations, are too simple for VQA. It is doubtful whether such synthetic images are representative enough to reflect the complexity of real-world images.

Recently, Hudson and Manning~\cite{hudson2019gqa} proposed a new dataset GQA for VQA, which provides scene graph annotations for real-world images. By utilising the scene graph annotations and further data cleaning, we contribute a new dataset named \sexyname for referring expression, which contains not only region-expression pairs with complex reasoning chains but also visually similar distractors. It demands a much stronger reasoning ability to understand the whole expression and distinguish subtle visual differences in the images. Note that GQA also provided experiments localising related regions for questions but it was only regarded as a metric to evaluate the VQA task rather than targeting at the REF task. Neither expressions or distractors were considered in their setting.

\begin{table*}[t]
\centering
\renewcommand{\arraystretch}{1.1}
\setlength{\tabcolsep}{0.15em}

\begin{tabular}{|c|l|l|l|l|}
\hline
Index & Forms  & Reasoning trees  & Exemplar templates  &Expression examples  \\
\hline
\hline
\multirow{2}{*}{1} & \multirow{2}{*}{chain}       &  {\small{obj$_0$ (att$_0$) $\xrightarrow{\text{rel}_0}$ obj$_1$ (att$_1$) }}  & \small{The $<$att$_0$$>$ $<$obj$_0$$>$ that is $<$rel$_0$$>$ }  &  \small{The young girl that is touching the }  \\
  &         & $\xrightarrow{\text{rel$_1$}}$ obj$_2$ (att$_2$) &  \small{$<$att$_1$$>$ $<$obj$_1$$>$ that is $<$rel$_1$$>$ $<$obj$_2$$>$.}   & \small{glazed donut that is on the round table.} \\
\hline
\multirow{2}{*}{2} & \multirow{2}{*}{and}           & \multirow{2}{*}{\footnotesize{obj$_0$ (att$_0$)} \scriptsize{$\begin{cases} \tiny{\xrightarrow{\text{\footnotesize{rel$_0$}}}} \footnotesize{\text{obj$_1$ (att$_1$)}} \\ \tiny{\xrightarrow{\text{\footnotesize{rel$_1$}}}} \footnotesize{\text{obj$_2$\ (att$_2$)}} \\\end{cases}  $}}    & \small{The $<$att$_0$$>$ $<$obj$_0$$>$ $<$rel$_0$$>$ the $<$att$_1$$>$ $<$obj$_1$$>$}  & \small{The white fence near the building}  \\
  &                    &    & \small{and $<$rel$_1$$>$ the $<$att$_2$$>$ $<$obj$_2$$>$.} & \small{and behind the walking woman.}  \\
 \hline
\multirow{2}{*}{3} &  \multirow{2}{*}{or}            & \multirow{2}{*}{\footnotesize{obj$_0$ (att$_0$)} \scriptsize{$\begin{cases} \tiny{\xrightarrow{\text{\footnotesize{rel$_0$}}} \footnotesize{{\text{obj$_1$ (att$_1$)}}}} \\  \tiny{\xrightarrow{\text{\footnotesize{rel$_1$}}} \footnotesize{\text{obj$_2$\ (att$_2$)}}} \\\end{cases}  $}} & \small{The $<$att$_0$$>$ $<$obj$_0$$>$ $<$rel$_0$$>$ the $<$att$_1$$>$ $<$obj$_1$$>$} &  \small{The green suitcase behind the black}  \\
  &              &   & \small{or $<$rel$_1$$>$ the $<$att$_2$$>$ $<$obj$_2$$>$.} & \small{suitcase or near the yellow suitcase.} \\
\hline
4 &  order             &  \small{obj$_0$ (idx, dir, att$_0$)} & \small{The $<$idx$>$ $<$obj$_0$$>$ from the $<$dir$>$ that is  $<$att$_0$$>$.}  & \small{The first glass from the left that is red.} \\
\hline
\multirow{2}{*}{5} &  \multirow{2}{*}{same}   & \multirow{2}{*}{\small{obj$_0$ $\xrightarrow{\text{same cat}}$ obj$_1$}} & \small{The $<$obj$_0$$>$ that has the same $<$cat$>$}   &  \small{The bag that has the same color as }\\
 &                    &     & \small{as the $<$obj$_1$$>$.}  & the sweater.\\
\hline
6 &  not    & obj$_0$ (not att$_0$) & \small{The $<$obj$_0$$>$ that is not $<$att$_0$$>$.}  & \small{The apple that is not red.}\\
\hline
\end{tabular}
\vspace{-1em}
\caption{Examples of expression logic forms. Attributes of the objects are bounded with $()$ and relations between objects are shown on $\xrightarrow{}$. obj$_0$ denotes the target object, while obj$_{1,2}$ denote the related objects. att$_{0,1,2}$ and rel$_{0,1,2}$ denote the corresponding attributes and relations, respectively.}\label{tab:form}
\vspace{-1.5em}
\end{table*}

\vspace{-1.5em}
\paragraph{Referring Expression Models.} Referring expression~\cite{Dogan_2019_CVPR, hu2017modeling,hu2016natural,kong2014you,mao2016generation,matuszek2012joint,sadeghi2011recognition,yang2019cross,yang2019fast,Ye_2019_CVPR} has attracted great attention. Karpathy and Fei-Fei~\cite{karpathy2015deep} learned visual alignments between text and regions by multiple instance learning. Rohrbach~\etal~\cite{rohrbach2016grounding} localised a region by reconstructing the sentence using an attention mechanism. \cite{yu2016modeling,nagaraja2016modeling,zhang2018grounding} utilised context information to ground the expression. Yu~\etal~\cite{yu2018mattnet} and Liu~\etal~\cite{liu2019learning}, respectively, used modular networks and neural module tree networks to match better structure semantics. Following~\cite{yu2018mattnet}, Wang~\etal~\cite{Wang_2019_CVPR} and Liu~\etal~\cite{liu2019improving} increased the reasoning ability by watching neighbour regions and cross-modal attention-guided erasing. Different from these previous studies focusing on referring short expressions in a single image, we refer complex expressions in multiple similar images, which is more challenging and requires a stronger visual reasoning ability.
\vspace{-1.4em}
\paragraph{Text based Image Retrieval.} Text based image retrieval returns relevant images from the gallery that is described by the text description~\cite{barnard2003matching,gong2014multi,lee2018stacked,li2019visual,lin2014microsoft,plummer2015flickr30k,socher2014grounded,wang2016structured,hexiang2018bison}. 
Different from text based image retrieval, \sexyname focuses on fine-grained region-level matching. The distracting regions in \sexyname are more semantically similar to the relevant region in the target image with only subtle differences. Such fine-grained and region-level similarity requires models with much stronger reasoning ability to ground the flowery expressions. 

\vspace{-0.5em}
\section{The \sexyname Dataset and Task}
\vspace{-0.5em}
Previous natural image referring expression datasets~\cite{kazemzadeh2014referitgame,mao2016generation,yu2016modeling} typically only require the ability to recognise objects, attributes and simple relations. Apart from such shallow ability, our proposed \sexyname also measures deeper reasoning ability like logic and relational inference. Compared with previous datasets, it has two main features, namely 1) flowery and compositional expressions which need complex reasoning ability to understand and 2) a challenging test setting that includes controlled distractors with similar visual properties to the referent. Fig.~\ref{fig:task} shows a typical example of our dataset. In the following subsections, we first introduce the construction of the dataset, including generating expressions (Sec.~\ref{sub:expEngine}), discovering distractors (Sec.~\ref{sub:DisDist}) and post-processing (Sec.~\ref{sub:post}). We then analyse the statistics of our dataset in Sec.~\ref{sub:statistics}. We formally define the task in Sec.~\ref{sub:task}.

\vspace{-0.4em}
\subsection{The Expression Engine}
\vspace{-0.5em}
\label{sub:expEngine}
The expression engine is the key to the construction of our dataset, responsible for generating grammatically correct, unambiguous and flowery expressions with various compositionality for each of the described regions. We propose to generate expressions from scene graphs based on some logic forms. Specifically, given a region to be described, we first choose a logic form from a predefined logic family and obtain a textual template for it. We then take the target object node in the scene graph as a root and expand it into a specific reasoning tree needed for the textual template. Finally, we fill the textual template with the content parsed from the reasoning tree and produce the expression. In the following paragraphs, we will describe the details of three steps.

\vspace{-1.5em}
\paragraph{Expression logic forms.}
Expression logic forms summarise the abstract logics and provide specific structures for the expressions. Each of them is associated with several textual templates. Specifically, we define six types of expression logic forms, namely \emph{chain}, \emph{and}, \emph{or}, \emph{order}, \emph{same}, and \emph{not}.
These high-level logic forms provide different contexts for the target object. Specifically, \emph{chain}, \emph{and} and \emph{or} describe the relationship between the target object and other related objects. The \emph{chain} form considers a sequence of related objects connected by a chain, while the \emph{and} form indicates the target object must have some specific relations with two other objects and the \emph{or} form only requires fulfilling one of the two relations. The \emph{order} form provides relative spatial location between the target object and other objects of the same category. The \emph{same} form shows that the target object shares the same attributes as the related object. The \emph{not} form indicates a certain attribute or relation being absent in the target object. These basic logic forms can be further composed with each other and generate more complex and compositional expressions. The logic forms and their templates are shown in table \ref{tab:form}.

Although these logic forms cannot fully reflect the complexity of natural language, the basic logic units covered and their flexible compositions are sufficient to evaluate a model's reasoning ability. Moreover, experimental results show that knowledge learned from the \sexyname dataset can be directly applied to previous human-annotated datasets like refCOCO.

\vspace{-1.7em}
\paragraph{Reasoning tree parsing.} While expression logic forms define the structures for the expressions, the dense scene graphs provide the corresponding semantic content. We use the scene graphs provided in \cite{hudson2019gqa,krishna2017visual} to represent the internal semantic structures of images. Each node represents an object and edges between nodes represent the relations between them.
Textual templates of different expression logic forms require different semantic content as input, which can be represented by reasoning trees with different structures extracted from the scene graph. Table~\ref{tab:form} shows an instantiation of the reasoning tree for each of the forms, their corresponding textual templates and expression examples.

Specifically, for the \emph{chain}, \emph{and} and \emph{or} forms, we simply parse the needed semantic reasoning trees from the scene graphs. For the \emph{order} form, we sort all the regions that are of the same object category from left to right (vice versa) based on the coordinates of the centres of the bounding boxes. Since the order constraint is rather weak (\eg, `\texttt{the left glass}' may also exist in the distracting images), we further add additional attributes and/or relations to make the expression unambiguous. Similarly, for the \emph{not} form, we traverse the whole scene graph and collect the attributes/relations that present in all objects of the same category but not in the target object. For the \emph{same} form, we find the attributes that only the target object and the related object have, and regard the category of the attribute as a relation between the two objects. The attribute categories used in the \emph{same} form include colour, shape, material, gender and pattern.
\vspace{-1.5em}
\paragraph{Expression Generation.} With the expression logic forms and the parsed reasoning trees ready, the expression engine can generate flexible expressions by filling the textual templates of the expression logic forms with the content from the reasoning tree.
For example, given the \emph{order} form and a textual template like \texttt{the <index> <object> from <direction>}, the expression engine can generate `\texttt{the first glass from the left}' for the reasoning tree, \textit{glass (first, left)}. It can also generate more flowery expressions by adding more attributes or nodes to the reasoning tree. For example, it can produce `\texttt{the first glass from the left that is clear and to the left of the gravy}' by the expanded reasoning tree, \textit{glass\ (first,\ \ left, \ clear)$\xrightarrow{\textit{to the left of}}$ gravy}.
\subsection{Discovery of Distracting Images}
\vspace{-0.5em}
\label{sub:DisDist}
Introducing distracting images in the testing phase is another important feature of our proposed dataset. It provides more complex visual reasoning context and reduces dataset bias. The inclusion of distracting images guarantees that good performance can only be achieved by REF models that are able to reason over the complete expression and distinguish subtle visual differences.
We define four types of distracting images, namely:
\vspace{-0.7em}
\begin{enumerate}
    \item \texttt{DiffCat}: images that contain objects of different categories as the target object. 
    \vspace{-0.7em}
    \item \texttt{Cat}: images that contain objects of the same category as the target object. 
    \vspace{-0.7em}
    \item \texttt{Cat\&attr}: images that contain objects of both the same category and attributes as the target object. 
    \vspace{-0.7em}
    \item \texttt{Cat\&cat}: images that contain all the objects in the reasoning tree but of different relations.
\end{enumerate}
These distractors can be used to evaluate different aspects of REF models such as object recognition, attribute identification and relation extraction, etc. They force the models to fully understand the flowery and compositional expressions to achieve good performance. For each expression in the validation set and testing set, we provide 3 distracting images under each distracting factor, apart from the image containing the ground-truth target. We simply discard those region-expression pairs for which we can not find enough distractors. Fig.~\ref{fig:task} shows an example of distracting images of different types for a given expression.

\begin{figure*}[t]
    \centering
    \begin{minipage}[t]{0.31\textwidth}
    \centering
    \includegraphics[width=\textwidth,height=0.46\textwidth]{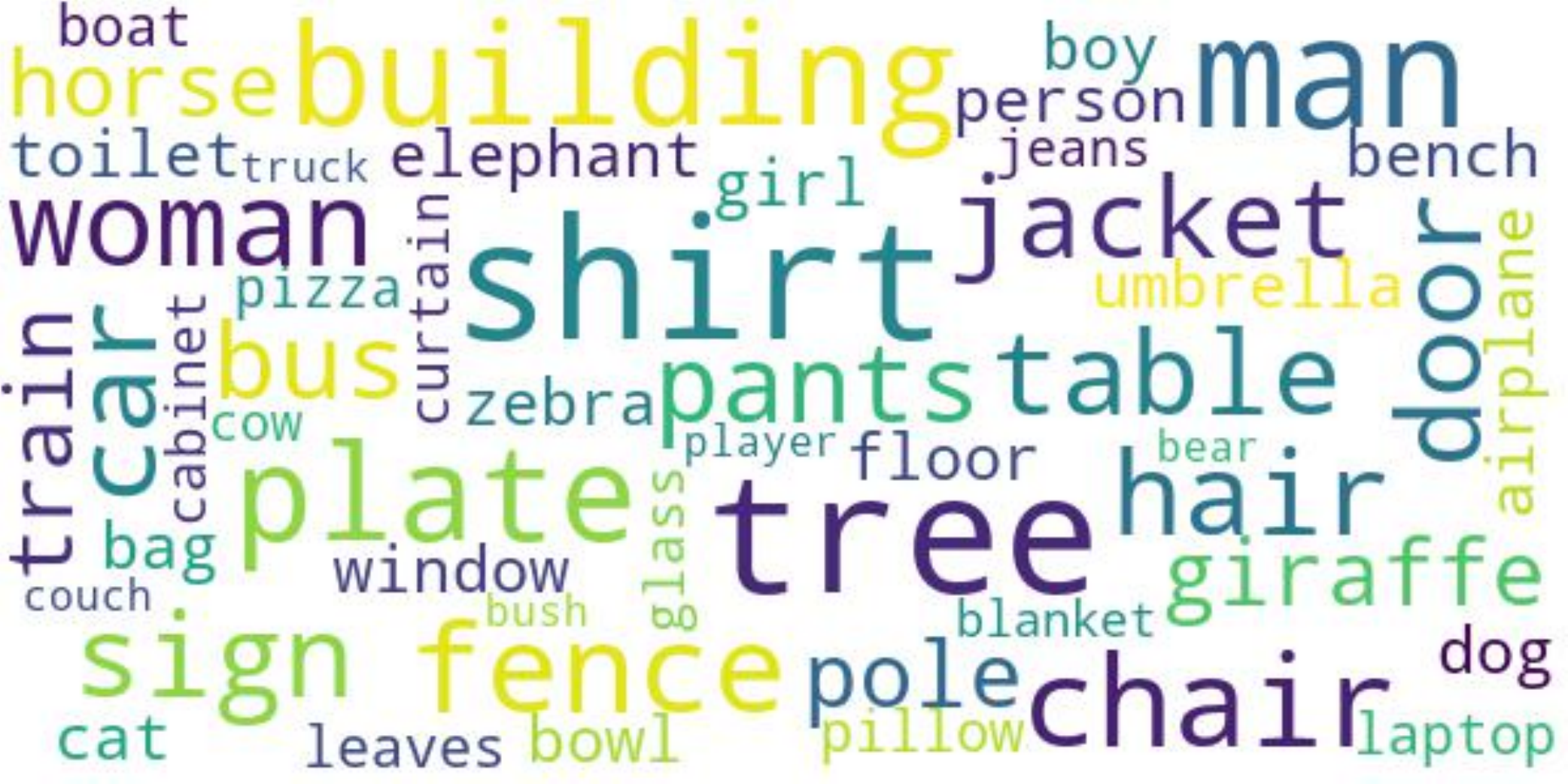}
    \small{(a) The most frequent object names}
     \vspace{-1em}
    \end{minipage}
    \begin{minipage}[t]{0.31\textwidth}
    \centering
    \includegraphics[width=\textwidth,height=0.46\textwidth]{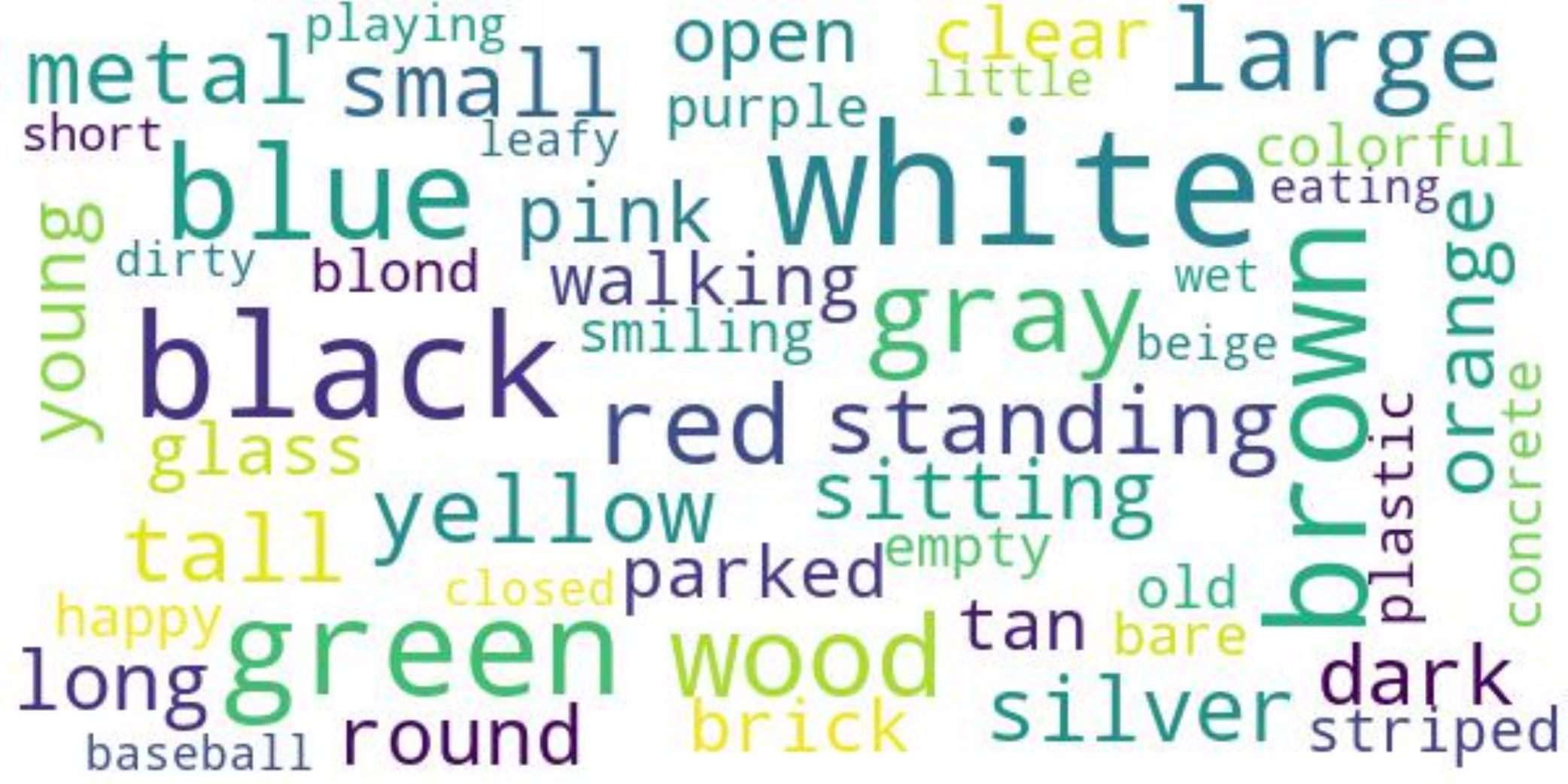}
    \small{(b) The most frequent attributes}
    \vspace{-1em}
    \end{minipage}
    \begin{minipage}[t]{0.31\textwidth}
    \centering
    \includegraphics[width=\textwidth,height=0.46\textwidth]{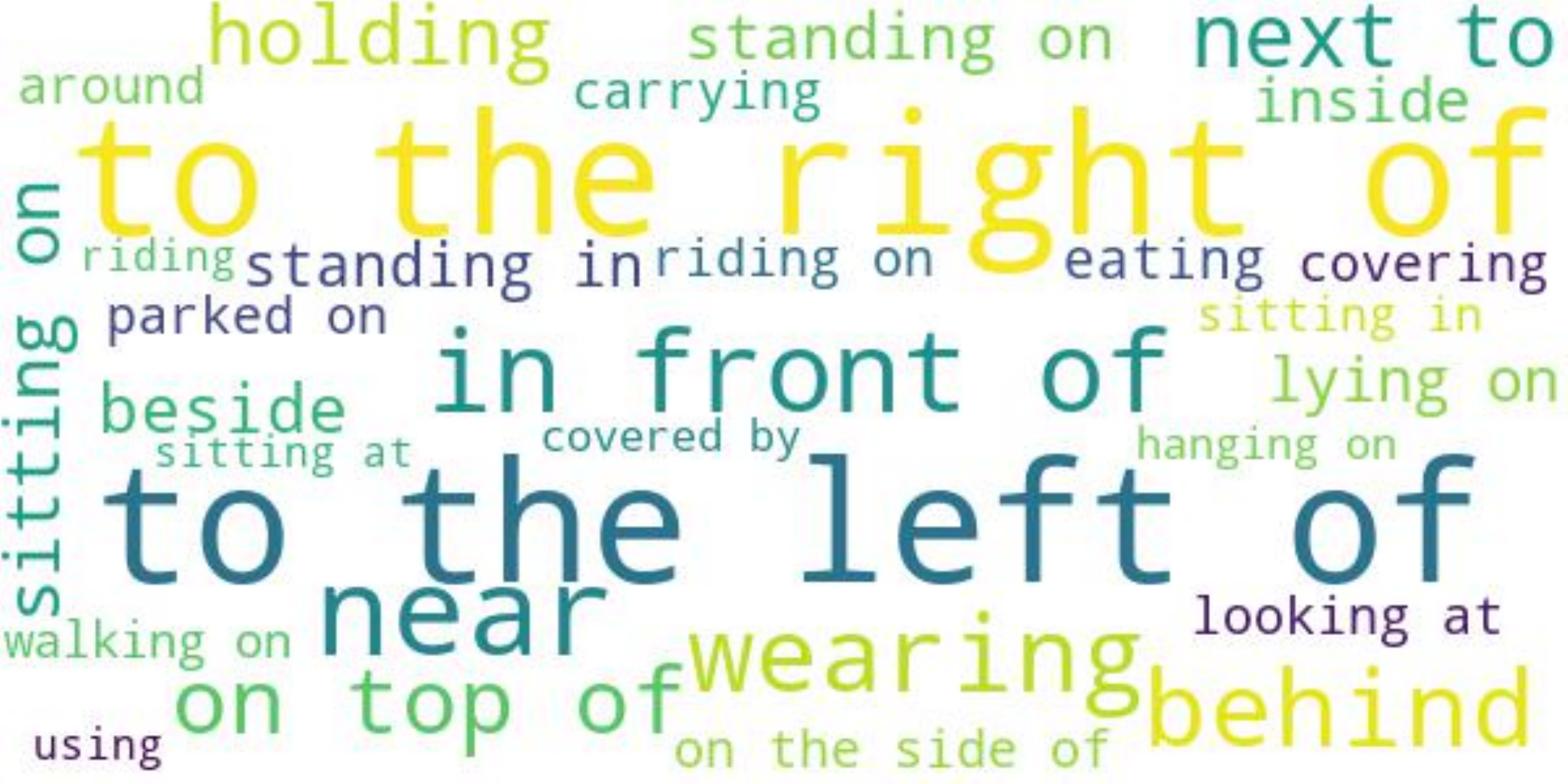}
    \small{(c) The most frequent relations}
    \vspace{-1em}
    \end{minipage}
    \caption{The most frequent object names, attributes and relations of \sexyname. The size of words indicates frequency.} \label{fig:vis}
    \vspace{-1em}
\end{figure*}

\subsection{Post Processing and Balancing}
\vspace{-0.5em}
\label{sub:post}
We use synonyms parsed from wordNet~\cite{Miller95wordnet:a} to further improve the diversity of the expressions. Besides, we remove expressions that target at classes that are hard to be bounded by a regular rectangle box (\eg, `\texttt{sky}' and `\texttt{cloud}') and regions that are too small (\ie, regions whose area is smaller than $1\%$ of the whole image). We also notice that some of the scene graph annotations in GQA are incorrect or incomplete ( \eg, missing annotations for some objects/attributes/relations). They may make some regions in the distracting images also semantically match the expressions. To avoid such noise in the distractors, we manually check the expressions and images in the testing set and discard these pairs with noise.

We also find some simple relations like `\texttt{to the left of}' being much more frequent than others in the scene graphs of GQA~\cite{hudson2019gqa}. To address such bias issues, we adopt two strategies: 1) we sample relations for each node based on a probability that is directly proportional to the reciprocal of the frequency, downsampling most frequent relations and enriching diversity of expressions; 2) we abandon those expression-regions that only contain simple spatial relations.

\vspace{-0.5em}
\subsection{Statistics of the Dataset}
\vspace{-0.5em}
\label{sub:statistics}
After the above processing and balancing, we have 148,712 expressions and 1,307,885 regions on 75,299 images, making our dataset the current largest real-world image dataset for referring expressions. The average length of the expressions is 14.4 and the size of the vocabulary is 1,596. Since the scene graph annotation of the testing set of GQA dataset is {\bf not} publicly released, we use the validation set of GQA to construct our testing set.
A new validation set is split out from the training data of GQA to monitor the model training process.
There are 119,603/16,524/12,586 expressions for training/validation/testing, respectively.

\begin{table}[t]
\centering
\renewcommand{\arraystretch}{0.9}
\setlength{\tabcolsep}{0.3em}
\resizebox{0.85\linewidth}{!}{
\begin{tabular}{|l|cccccc|}
\hline
             & Object  & Att. &Rel.  & Exp.   &  Cand & Cat Cand   \\
             & Cat.    & Num. &Num.  & length &  Num. &  Num.      \\
 \hline
 \hline
 \small{refCOCO}    & 80{\color{red}{$^{1}$}} & - & - & 3.5 & 10.6  & 4.9  \\
 \small{refCOCOg}    & {80} & - & - & 8.5 & 8.2  & 2.6  \\
 \footnotesize{CLEVR-Ref+}  &  3  &  12   & 5&  22.4 & -  & - \\
 \small{\sexyname}   & 508  & 601 & 299 & 14.4  & 262.5 & 20.3 \\ \hline
\end{tabular}}
\vspace{-1em}
\caption{Statistic comparison of refCOCO~\cite{yu2016modeling}, refCOCOg~\cite{mao2016generation}, ClEVR-Ref+~\cite{liu2019clevr} and the \sexyname on number of object categories, number of attributes, number of relations, average length of expressions, average number of object candidates and average number of object candidates that are of the same categories for each expression.} \label{tb:cmp}
\vspace{-1.5em}
\end{table}

Thanks to the dense annotations of the scene graphs, the proposed dataset contains fine-grained annotations for object categories, attributes and relations. The number of entry-level object categories, attributes and relations are 508, 601, and 299, respectively. We show the most frequent object names, attributes and relations in Fig.~\ref{fig:vis}. We can see diverse object categories, with `\texttt{man}', `\texttt{building}' and `\texttt{tree}' being the most frequent object names. The most frequent attributes are colours (~\eg `\texttt{black}' and `\texttt{white}') and sizes (~\eg `\texttt{small}' and `\texttt{large}') while the most frequent relations are spatial relations like~`\texttt{to the left/right of}'.  We compare the statistics of the proposed \sexyname dataset with three widely-used referring dataset, refCOCO~\cite{yu2016modeling}, refCOCOg~\cite{mao2016generation} and CLEVR-Ref+~\cite{liu2019clevr} in table~\ref{tb:cmp}\footnote{The definition of object categories between \sexyname and refCOCO are of different hierarchies. \sexyname does contain more diverse object categories like `tree', `shirt' and `mat' which don't exist in refCOCO.}. 
As shown in table~\ref{tb:cmp}, the proposed dataset enjoys diverse object categories, attributes and relations. Moreover, it provides reasonably long expressions and much more candidate objects of same/different categories as the target object. The average length of our expressions is shorter than that of CLEVR-Ref+, but we find it is not necessary to use longer expressions to distinguish the target object in the real-world images even when distractors exist.
More analysis about dataset bias and baseline results can be found in Sec.~\ref{sec:exp} and we provide more data examples and detailed statistics in the supplementary material.

\subsection{Task}
\vspace{-0.5em}
\label{sub:task}
Given a natural language referring expression and a set of similar images, the proposed \sexyname task requires a model to localise a target region described by the expression. Compared to the previous REF task~\cite{kazemzadeh2014referitgame,mao2016generation}, the \sexyname demands a better understanding of longer and flowerier expressions, and the ability to distinguish the subtle differences of the distracting images. It requires REF models to have stronger reasoning ability for object detection, attribute recognition and relation extraction. Formally, given $N$ images and a query expression $q$, the \sexyname task identifies a target region $r_{i^*, j^*}$ by
\vspace{-0.5em}
\begin{equation}
    \small
    r_{i^*,j^*} = \mathop{\arg\max}_{r_{i,j}, i \in [1, N], j \in [1, J_i]}s(r_{i,j}|q),
\end{equation}
where $I_i$ denotes the $i$-th image, $r_{i,j}$ is the $j$-th region from $I_i$, $J_i$ is the number of the regions in $I_i$, $s(r_{i,j}|q)$ denotes the matching score between $r_{i,j}$ and $q$.

Note that we do not provide distracting images during training because they are usually unavailable or hard-to-collect in the real world. {Also, it is easier for us to follow the original training strategies in~\cite{lee2018stacked,liu2019improving,yu2018mattnet,rohrbach2016grounding} to re-train and evaluate the models.}


\vspace{-0.5em}
\section{Model}
\vspace{-0.5em}
Although \sexyname is a new task that requires localising a region from a set of images instead of a single one,  existing REF models can be directly applied to this new task by densely matching the query expression with each object in the image set and choosing the one with the highest matching score as the referring result.

MattNet~\cite{yu2018mattnet} is a popular backbone model for solving the REF task because of its extraordinary capability in modeling different modules of the query expressions, including subject ($sub$), location ($loc$) and relationship ($rel$). Specifically, MattNet estimates the matching score between an expression $q$ and the $j$-th region $r_j$ by
\vspace{-0.5em}
\begin{equation}
    \small{
    s(r_j|q) = \sum_{md}w^{md}s(r_j|q^{md}),
    }
\vspace{-0.5em}
\end{equation}
where $ md \in \{sub, loc, rel\}$, $w^{md}$ is the learnt weight for the $md$ module and $q^{md}$ is the modular phrase embedding. More details about MattNet can be found in \cite{yu2018mattnet}.

Given a positive pair $(r_m, q_m)$, the whole model of MattNet is optimised by a ranking loss, given by
\vspace{-0.5em}
\begin{equation}
    \small{
    \begin{aligned}
    \mathcal{L}_{rank}=& \sum_{m}([\Delta-s(r_m|q_m)+s(r_m|q_n)]_{+}+\\
    &[\Delta-s(r_m|q_m)+s(r_o|q_m)]_{+}),
    \end{aligned}
    }
    \label{loss_rank}
    \vspace{-0.5em}
\end{equation}
where $r_o$ and $q_n$ are other random unaligned regions and expressions in the same image as $r_m$ and $q_m$, $\Delta$ is a margin and $[x]_+=max(x, 0)$.
This loss function is suitable for the REF task and can successfully distinguish aligned region-expression pairs from unaligned ones within the same image. However, when it comes to the \sexyname task, it has the limitation on not being able to identify hard negative examples with similar visual properties in other images, because the training of MattNet does not consider hard negative regions and expressions in other images. To solve this problem, we propose a modular hard-mining training strategy based on MattNet.

\vspace{-1.2em}
\paragraph{Modular Hard-mining Strategy.}
To increase the ability of MattNet to distinguish hard negative regions in distracting images, we need to sample distracting regions/expressions in other images as negative training examples. However, since there are 119,603 expressions and 797,595 regions in the training set of \sexyname, how to mine hard negative regions and expressions effectively and efficiently becomes a challenge. To handle this challenge, we propose to use the similarity of modular phrase embedding $q^{md}$ as a prior to sample the hard negative examples in other images, where $md \in \{sub, loc, rel\}$.

Specifically, for the $m$-th region-expression pair, we first extract its modular expression features $\{q^{md}_m\}$ and calculate their similarity with those of the $n$-th region-expression pair that has the same object category. We define the probability of sampling the $n$-th region-expression pair to be the negative pair by
\vspace{-0.5em}
\begin{equation}
    \small{
    \begin{aligned}
    s^{md}_{m,n} & = f(q^{md}_m, q^{md}_n), \\
    p^{md}_{m,n} & = \frac{ exp(s^{md}_{m,n})}{\sum_{n=1, m \neq n}^{n=N_C} exp(s^{md}_{m,n})},
    \end{aligned}
    }
    \vspace{-0.5em}
\end{equation}
where $f$ is a function for estimating the similarity between two expression features and $N_C$ is the number of the region-expression pairs has the same object category as the $m$-th region-expression pair in the training set. For simplicity, We use cosine similarity as an instantiation of $f$. We mine hard distracting regions and expressions for each positive region-expression pair and send these distracting regions to a ranking loss as hard negative examples.

Formally, our modular hard-mining loss is
\vspace{-0.5em}
\begin{equation}
    \small{
    \begin{aligned}
    \mathcal{L}_{mine}=& \sum_{m}\sum_{md}([\Delta-s(r_m|q_m)+s(r_m|q_n^{md})]_{+}+\\
    &[\Delta-s(r_m|q_m)+s(r_n^{md}|q_m)]_{+}),
    \end{aligned}
    }
    \label{loss_mine}
    \vspace{-0.5em}
\end{equation}
where $r_n^{md}$ and $q_n^{md}$ are a region-expression pair sampled with $\{p^{md}_{m,n}\}^{N_C}_{n=1, m \neq n}$ as a prior.

Our total loss is $\mathcal{L}=\mathcal{L}_{rank}+\mathcal{L}_{mine}$, where $\mathcal{L}_{rank}$ targets at distinguishing distracting negative regions and expressions within the same image, and $\mathcal{L}_{mine}$ targets at distinguishing similar negative regions and expressions in other images.

Such a modular hard mining strategy is effective since it can mine hard negative region-expression pairs outside the image that contains the target region-expression pair. Besides, the mined regions have similar properties as the target, which demand stronger reasoning ability to distinguish.
It is also efficient since it only requires the expressions as input without the need for loading images into memory. It enables the model to scan all the expressions in the training set in around 29 seconds with a na\"ive implementation. During training, we update the sample probability $p_{i,j}^{md}$ every 50 iterations. We distinguish the proposed hard mining model from the original MattNet by calling it \textbf{MattNet-Mine}.

\vspace{-0.5em}
\section{Experiments}
\vspace{-0.5em}
\label{sec:exp}
In this section, we conduct extensive experiments to analyse the \sexyname dataset and compare our proposed model with SOTA REF models. We first study the bias impact and transfer performance. We then compare the performance of the proposed MattNet-Mine with the baselines. We also provide extensive analysis, including ``retrieve'' + ``REF'' to handle the task, performance against logic forms and lengths of the expressions. We finally provide an ablation study on our mining strategy for distractors. We introduce experimental settings before we start.
\vspace{-0.5em}
\subsection{Experimental Settings}
\vspace{-0.5em}
\paragraph{Implementation Details.} Following MattNet~\cite{yu2018mattnet} and CM-Att-Erase~\cite{liu2019improving}, we extract visual features by res101-based Faster-RCNN~\cite{he2016deep,ren2015faster} pre-trained on COCO~\cite{lin2014microsoft}. For each word in the sentence, we initialise it with an one-hot word embedding. We train all the models with Adam optimiser~\cite{kingma2014adam} until the accuracy of the validation set stops improving. {We set the maximum time step for the text encoder to be 30. Expressions with words less than 30 are padded}. For other settings for hyper-parameters, we keep them the same as the original MattNet to avoid cumbersome parameter fine-tuning. For the proposed MattNet-Mine, we first pre-train it by the ranking loss $\mathcal{L}_{rank}$ to obtain reasonable modular attentive phrase embeddings and then finetune the model with both $\mathcal{L}_{mine}$ and $\mathcal{L}_{rank}$.
Following previous REF models like~\cite{liu2019improving,Wang_2019_CVPR,yu2018mattnet}, we use ground-truth object bounding boxes as proposals. We consider it as a correct comprehension if the model successfully chooses the proposal pointed by the expression among all the proposals extracted from the similar image set.

\begin{table*}[t]
    \centering
    {\small
    {
    \begin{tabular}{|l|c|cccc|c|}
    \hline
    Method                                                & Full &  DiffCat & Cat & Cat\&attr &  Cat\&cat & WithoutDist \\
    \hline
    \hline
    Chance                                                & 0.4 & 1.7  &  1.8 & 1.9          &  1.7    & 6.6 \\
    GroundeR~\cite{rohrbach2016grounding}                 & 19.1     & 60.2    &  38.5 & 35.7  & 38.9  & 75.7 \\
    Deaf-GroundeR                                      & 2.2  & 7.7  &  7.9      & 8.0       & 8.0  &   27.1  \\
    Shuffle-GroundeR                                  & 13.1 & 41.8 & 28.6      & 27.2      & 27.6 &   58.5 \\
    Obj-Attr-GroundeR                                 & 15.2 & 53.1 & 32.6      & 29.6      & 32.7 &   68.8\\
    MattNet-refCOCO       & 8.7  &   22.7&  17.0      &     16.7   & 18.9  &   42.4\\
    \hline
    \hline
    MattNet~\cite{yu2018mattnet}                          & 26.3 & 69.1 & 45.2      & 42.5      & 45.8 & 77.9 \\
    CM-Att-Erase~\cite{liu2019improving}                  & 28.0 & \textbf{71.3} & 47.1      & 43.4      & 48.4 & \textbf{80.4} \\
    SCAN \cite{lee2018stacked}+MattNet                     & 18.8&-&-&-&-&-\\
    \hline
    \hline
    MattNet-Mine                                          & \textbf{33.8} & 70.5 & \textbf{54.4}      & \textbf{46.8}      & \textbf{52.0} & 78.4 \\
    \hline
    \end{tabular}}}
    \vspace{-1em}
    \caption{Results of baselines and state-of-the-art models on the \sexyname dataset. MattNet-refCOCO is trained on refCOCO.}
    \vspace{-1.5em}
    \label{tab:ref}
\end{table*}

\vspace{-1.5em}
\paragraph{Evaluation Settings.}
Table~\ref{tab:ref} shows different experiments settings. \texttt{Full} denotes the case when all the distractors are added while \texttt{WithoutDist} denotes no distractor is added. \texttt{DiffCat}, \texttt{Cat} and \texttt{Cat\&attr}, respectively, represent the cases when certain type of distractors are added, including distracting images containing no object of the same category as the target object, images containing objects of the same category, images containing objects of the same category and attributes and images contain all the objects in the reasoning tree but of different relations.

\vspace{-0.5em}
\subsection{Dataset Analysis}
\label{exp_data}
\vspace{-0.5em}
\paragraph{Bias Analysis.}
Inspired by the bias analysis of Cirik~\etal~\cite{cirik2018visual}, we use similar ways to analyse the bias problem of \sexyname. To exclude the impact of specific models or mechanisms, we choose GroundeR~\cite{rohrbach2016grounding}, which is the simplest CNN-RNN baseline model for referring expression.
We train several variants of GroundeR models, include deaf-GroundeR which masks out the language input of the GroundeR with an all-zero vector, shuffle-GroundeR which shuffles the order of the word sequence in the expression and Obj-Att-GroundeR which only keeps the nouns and adjectives of the text input.

The upper section of table~\ref{tab:ref} shows the results of the bias experiments. Deaf-GroundeR, an image only model achieves better performance than the ``Chance'' model, which selects a region from the images by chance. We observe that Deaf-GroundeR can filter out some irrelevant regions by providing higher matching scores for those regions whose categories like `woman' and `shirt' frequently appear in both the training set and testing set. This indicates that the statistics bias problem in previous datasets like refCOCOg~\cite{mao2016generation} also exists in our dataset. However, comparing the results of \texttt{WithoutDist} and \texttt{Full}, we see that the biased performance becomes much lower when distractors are added. Moreover, the bias problem in \sexyname is less significant than in refCOCOg. Deaf-GroundeR only achieves an accuracy of 2.2 in the \texttt{Full} case, while a similar ``image only'' model achieves an accuracy of 40.1 in \cite{cirik2018visual}.

Cirik~\etal~\cite{cirik2018visual} also pointed out that shuffling the order of expressions and masking out other words that are not nouns or adjectives have minor effect on the performance of refCOCOg, resulting in only a relative drop of 4\% and 3\%, respectively. This suggests that a model does not need very strong reasoning ability for the whole sentence to handle the task. However, in \sexyname, comparing Shuffle-GroundeR and Obj-Att-GroundeR with GroundeR under the \texttt{Full} case, we observe a relative drop of 31\% and 20\%, respectively. It indicates that the syntactic structure and relations play more significant roles in \sexyname regarding performance improvement.

\vspace{-1.5em}
\paragraph{Transfer Performance.}
We directly apply a MattNet~\cite{yu2018mattnet} model trained on refCOCO to our \sexyname and it only achieves an accuracy of 42.4 and 8.7 under the \texttt{WithoutDist} and \texttt{Full} cases, respectively. It shows our dataset and task are more complex and challenging. 
In contrast, a MattNet trained on \sexyname can achieve an accuracy of 56.5 and 64.5 on the testA and testB splits of refCOCO, which are about 65.7\% and 76.4\% of the performance of the original model trained on refCOCO, respectively.
This demonstrates the realism of our synthetic expressions and that the knowledge learnt from \sexyname can be directly transferred to real datasets like refCOCO, while the reasoning ability gained from refCOCO cannot solve our \sexyname task.

\vspace{-0.5em}
\subsection{Overall Evaluation}
\vspace{-0.5em}
We evaluate the proposed \sexyname task with three baselines, namely GroundeR~\cite{rohrbach2016grounding},  MattNet~\cite{yu2018mattnet} and CM-Att-Erase~\cite{liu2019improving}. GroundeR is a simple CNN-RNN baseline. MattNet is one of the most popular REF models, and CM-Att-Erase was the best state-of-art in REF at the time of this submission. We densely ground the expressions on every image in the similar image set and choose the region with the
highest score as the grounding result.

\vspace{-1.5em}
\paragraph{Performance of the REF Models.}
Table~\ref{tab:ref} reports the accuracy of all the baselines and the proposed MattNet-Mine.
We have the following observations.
(1) The performance gradually increases from GroundeR~\cite{rohrbach2016grounding} to MattNet~\cite{yu2018mattnet} and from MattNet~\cite{yu2018mattnet} to CM-Att-Erase \cite{liu2019improving}. This is consistent with their performance on the refCOCO, refCOCO+ and refCOCOg~\cite{kazemzadeh2014referitgame,mao2016generation}.
(2) The performance of these REF models decreases dramatically when distracting images containing the objects of the same object category are added, especially under the \texttt{Full} case. Among the 4 types of distractors, \texttt{DiffCat} affects the performance least while \texttt{Cat\&Attr} affects most. This implies that existing REF models strongly rely on the object and attribute recognition to localise the target region.
(3) Comparing with the original MattNet~\cite{yu2018mattnet}, our MattNet-Mine show improved performance under all cases, especially for the cases that contain fine-grained similar distractors. This demonstrates the effectiveness of the proposed hard mining strategy.
\vspace{-1.5em}
\paragraph{Performance of ``Retrieve'' + ``REF'' Strategy.}
We also evaluate another strategy to solve the problem in which we first use a text-based image retrieval model to select one image with the highest matching score and then ground the query expression in the selected image. We use SCAN (t2i)~\cite{lee2018stacked} as the retrieval model for its excellent performance, 
and we use MattNet to ground the expression in the returned image. We achieve an accuracy of 18.8 under the \texttt{Full} case. Compared with the other models in table~\ref{tab:ref}, the ``Retrieve''+``REF'' strategy performs worse than densely referring the query expression in every image. We believe this may be caused by the fact that densely referring an expression in every image provides more fine-grained regional level matching than the retrieval model.

\begin{figure}[t]
    \centering
    \includegraphics[width=0.8\linewidth,height=0.33\linewidth]{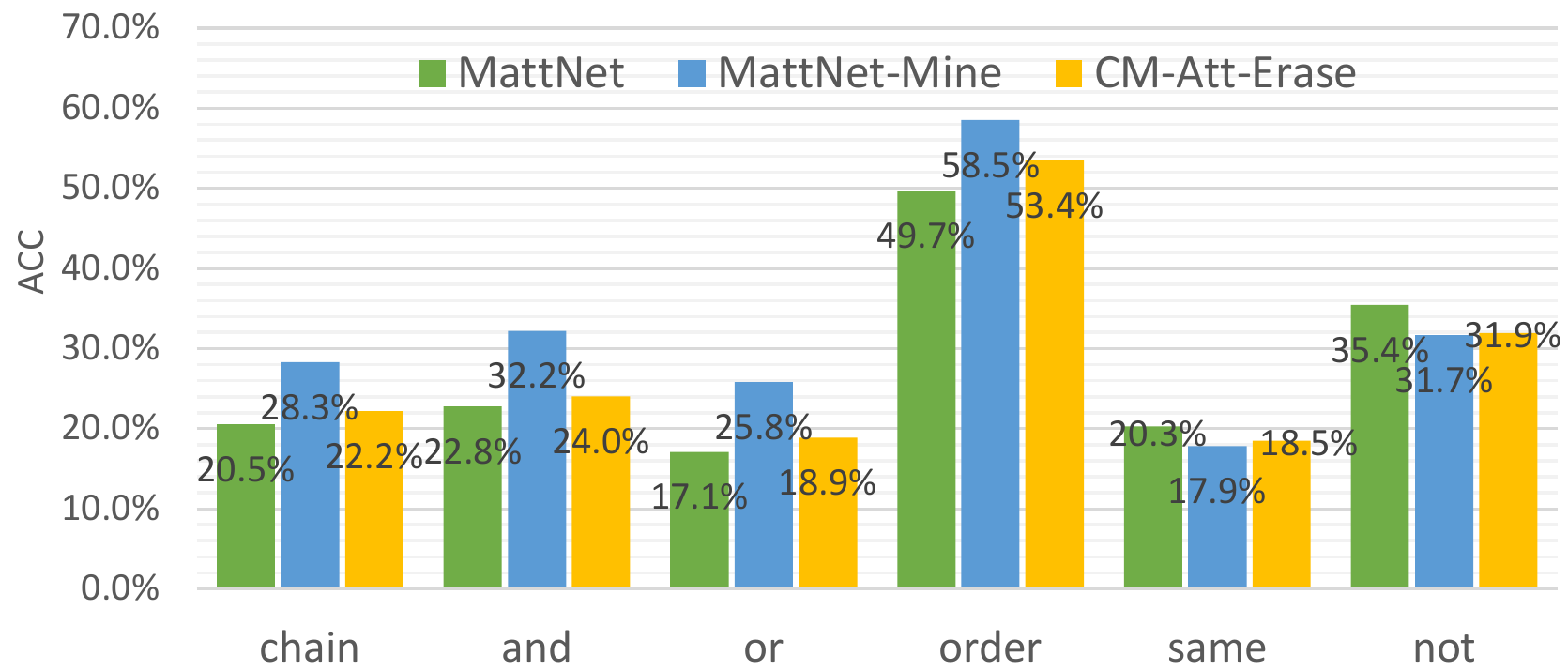}
    \vspace{-1em}
    \caption{Accuracy of expressions of different logic form.}
    \vspace{-0.5em}
    \label{fig:exp_form_logic}
    \vspace{-1em}
\end{figure}
\vspace{-1.6em}
\paragraph{Performance of Different Logic Forms.} We show the performance of expressions of each logic form in Fig.~\ref{fig:exp_form_logic}. We can see that while expressions of \emph{chain}, \emph{and}, \emph{or} and \emph{same} forms have similar accuracy, \emph{order} and \emph{not} forms have the best and second best accuracy, respectively. We believe the reasons are 1) the reasoning logic trees of \emph{order} and \emph{not} forms are simpler than other forms like \emph{chain}, \emph{and} and \emph{or} (see table~\ref{tab:form}), and 2) \emph{order} form has provided specific relative spatial location between the target object and the related objects of the same category within the same image.
\begin{figure}[t]
    \centering
    \includegraphics[width=0.68\linewidth,height=0.29\linewidth]{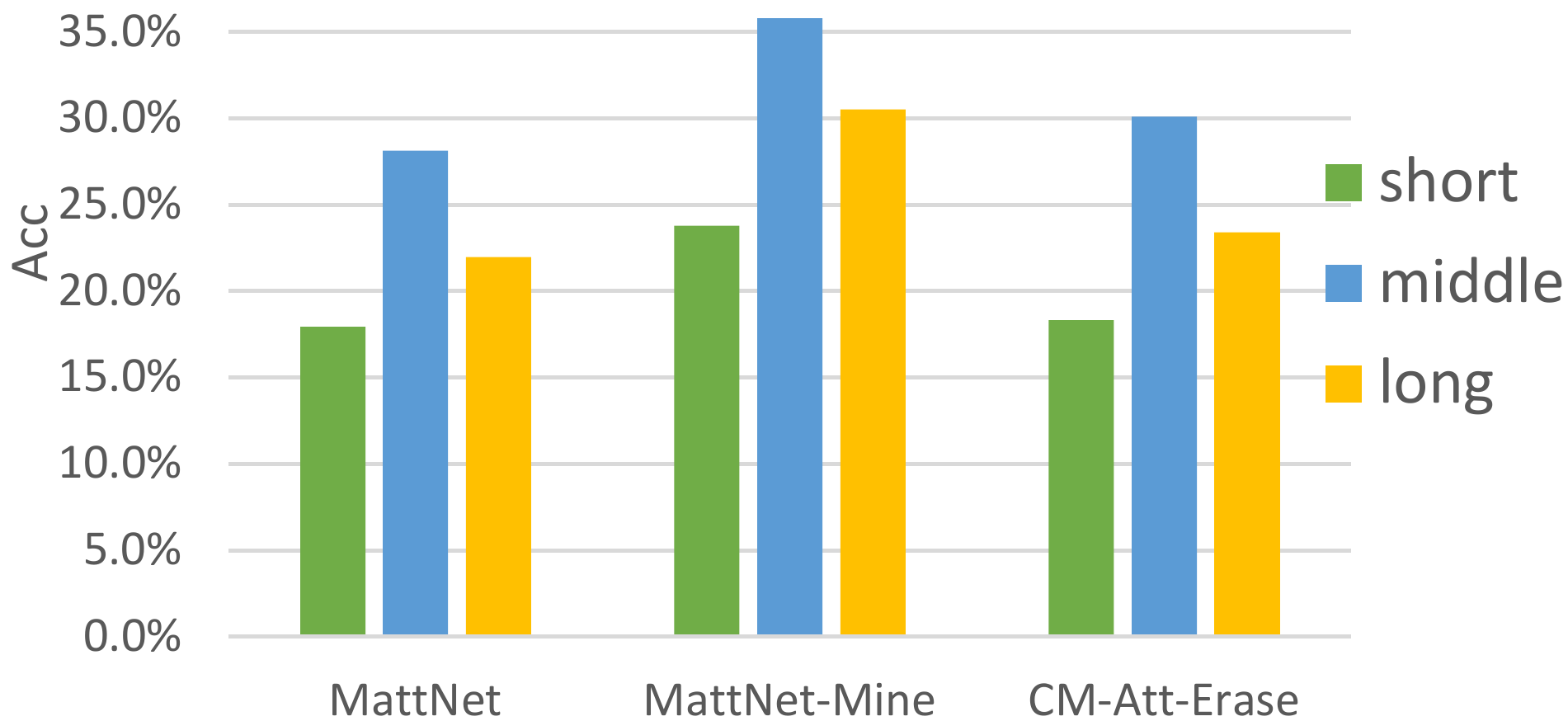}
    \vspace{-1em}
    \caption{Accuracy of expressions of different lengths.}
    \vspace{-2em}
    \label{fig:exp_form}
\end{figure}

\vspace{-1.5em}
\paragraph{Performance of Different Lengths of Expressions.} We divide expressions into 3 kinds based on the number of the words in the expressions, namely short (less than 10 words), middle (10-20 words) and long (more than 20 words), and test them separately. As shown in Fig. \ref{fig:exp_form}, we find that expressions in the middle group have the best accuracy. We suspect the reason is that short expressions provide limited textual information for distinguishing distracting regions while long expressions usually contain complex logic or semantics that requires stronger reasoning ability.
\begin{table}[t]
    \renewcommand{\arraystretch}{1}
    \setlength{\tabcolsep}{0.3em}
    \centering
    \resizebox{0.85\linewidth}{!}{
    \begin{tabular}{|l|ccccc|}
    \hline
    Method                                                & Full &  DiffCat & Cat & Cat\&attr &  Cat\&cat \\
    \hline
    \hline
    MattNet                                              & 26.3 &  69.1    & 45.2  & 42.5     & 45.8   \\
    Random                                               & 27.6 &  \textbf{71.6}    & 47.4  & 43.5     & 47.3   \\
    Class-aware                                                & 32.2 &  70.3    & 53.2  & 46.1     & 51.4  \\
    Sentence-sim                                               & 32.3 &  70.4    & 53.6  & 46.4     & 51.2  \\
    Module-specific                                                 & \textbf{33.8} & 70.5    & \textbf{54.4}  & \textbf{46.8}     & \textbf{52.0}  \\
    \hline
    \end{tabular} }
    \vspace{-1em}
    \caption{Ablation study of different hard mining strategies.}
     \label{tab:abs}
    \vspace{-1.5em}
\end{table}
\vspace{-0.5em}
\subsection{Ablation Study on Distractor Mining}
\vspace{-0.5em}
We conduct an ablation study to investigate different hard negative mining strategies for the \sexyname task. Specifically, we have the following solutions by replacing the $q_n^{md}$ and $r_n^{md}$ in Eq.~\ref{loss_mine} with features from different regions and expressions. ``Random'' means using regions and expressions that are randomly-selected from the whole dataset regardless the object category. ``Class-aware'' means using random-selected regions and expressions that has the same object category as the target region. ``Sentence-sim'' means a region-expression pair that are sampled based on the similarity of global expression features. We define the global expression features as the average embedding of all the words in the expression. ``Module-Specific" means the proposed modular specific hard mining strategy based on the similarity of the modular expression features.

Table~\ref{tab:abs} shows the ablation study results. Compared with the original MattNet, ``Random'' can provide an improvement under all cases. However, its improvement for the \texttt{Full} case is minor comparing with other mining strategy since it doesn't consider the similar distractors. ``Class-awre'' boosts the performance under the case where similar distractors are added, indicating the value of the distracting regions and expressions of the same category. ``Sentence-sim'' achieves only a comparable performance with ``Class-aware'', showing its inefficiency for hard negative mining. ``Module-specific'' achieves the best performance when similar distractors are added, showing its effectiveness to mine negative examples and distinguish similar distractors.

\vspace{-1.5em}
\section{Conclusion}
\vspace{-0.5em}
Expressions in existing referring expression datasets normally describe some simple distinguishing properties of the object which can not fully evaluate a model's visual reasoning ability.
In this paper, we proposed a new challenging dataset, \sexyname, for referring expression comprehension. The new dataset covers various reasoning logics that can be flexibly combined with rich visual properties. Additionally, to better exploit the full reasoning chain embodied in the expression, we proposed a new test setting by adding some additional distracting images. This newly proposed dataset suffers less bias and we found existing state-of-the-art models fail to show promising results. We then proposed a modular based hard mining strategy that achieves the best performance but is still far from perfect. We wish this new \sexyname dataset and task can attract more research attention and become a new benchmark in this area.

{\small
\bibliographystyle{ieee_fullname}
\bibliography{egbib}
}
\end{document}